\documentclass{article}

\usepackage{arxiv}

\usepackage[utf8]{inputenc} % allow utf-8 input
\usepackage[T1]{fontenc}    % use 8-bit T1 fonts
\usepackage{hyperref}       % hyperlinks
\usepackage{url}            % simple URL typesetting
\usepackage{booktabs}       % professional-quality tables
\usepackage{amsfonts}       % blackboard math symbols
\usepackage{nicefrac}       % compact symbols for 1/2, etc.
\usepackage{microtype}      % microtypography
\usepackage{lipsum}
\usepackage{graphicx}
\usepackage{comment}
\usepackage{multirow}
\usepackage{adjustbox}

\usepackage{subcaption}
\usepackage{amsmath}
\usepackage[psamsfonts]{amssymb}
\usepackage{amsxtra}
\usepackage{threeparttable}
\title{On Supervised Feature Selection from High Dimensional Feature Spaces}

\author{
  Yijing Yang \\
  University of Southern California\\
  Los Angeles, California, USA \\
  \texttt{yijingya@usc.edu} \\
     \And
  Wei Wang \\
  University of Southern California\\
  Los Angeles, California, USA \\
  \texttt{wang890@usc.edu} \\
   \And
  Hongyu Fu \\
  University of Southern California\\
  Los Angeles, California, USA \\
  \texttt{hongyufu@usc.edu} \\
   \And
  C.-C. Jay Kuo \\
  University of Southern California\\
  Los Angeles, California, USA \\
  \texttt{cckuo@sipi.usc.edu} \\
}

\begin{document}
\maketitle

\begin{abstract}

The application of machine learning to image and video data often yields
a high dimensional feature space. Effective feature selection techniques
identify a discriminant feature subspace that lowers computational and
modeling costs with little performance degradation. A novel supervised
feature selection methodology is proposed for machine learning decisions
in this work.  The resulting tests are called the discriminant feature
test (DFT) and the relevant feature test (RFT) for the classification
and regression problems, respectively. The DFT and RFT procedures are
described in detail. Furthermore, we compare the effectiveness of DFT
and RFT with several classic feature selection methods. To this end, we
use deep features obtained by LeNet-5 for MNIST and Fashion-MNIST
datasets as illustrative examples. Other datasets with handcrafted
and gene expressions features are also included for performance
evaluation. It is shown by experimental results that DFT and RFT can
select a lower dimensional feature subspace distinctly and robustly
while maintaining high decision performance. 

\end{abstract}

% \keywords{Authors should not add keywords, as these will be chosen during the submission process (please refer to Section II (F) below for further details).}

\maketitle

\section{Introduction}\label{sec:introduction}

Traditional machine learning algorithms are susceptible to the curse of
feature dimensionality \cite{hammer1962adaptive}. Their computational
complexity increases with high dimensional features. Redundant features
may not be helpful in discriminating classes or reducing regression
error, and they should be removed. Sometimes, redundant features may
even produce negative effects as their number grows. their detrimental
impact should be minimized or controlled. To deal with these problems,
feature selection techniques \cite{tang2014feature, miao2016survey,
venkatesh2019review} are commonly applied as a data pre-processing step
or part of the data analysis to simplify the complexity of the model.
Feature selection techniques involve the identification of a subspace of
discriminant features from the input, which describe the input data
efficiently, reduce effects from noise or irrelevant features, and
provide good prediction results~\cite{guyon2003introduction}. 

For machine learning with image/video data, the deep learning
technology, which adopts a pre-defined network architecture and
optimizes the network parameters using an end-to-end optimization
procedure, is dominating nowadays. Yet, an alternative that returns to
the traditional pattern recognition paradigm based on feature extraction
and classification two modules in cascade has also been studied, e.g.,
\cite{chen2018saak, chen2020pixelhop, chen2020pixelhop++, kuo2018data,
kuo2019interpretable, liu2021voxelhop, manimaran2020visualization,
rouhsedaghat2020facehop, zhang2020pointhop, zhang2020pointhop++}.  The
feature extraction module contains two steps: unsupervised
representation learning and supervised feature selection. Examples of
unsupervised representation learning include multi-stage Saab
\cite{kuo2019interpretable} and Saak transforms \cite{chen2018saak}.
Here, we focus on the second step; namely, supervised feature selection
from a high dimensional feature space. 

Inspired by information theory and the decision tree, a novel
supervised feature selection method is proposed in this work. The
resulting tests are called the discriminant feature test (DFT) and the
relevant feature test (RFT), respectively, for the classification and
regression problems. The DFT and RFT procedures are described in detail.
We compare the effectiveness of DFT and RFT with several classic feature
selection methods. Experimental results show that DFT and RFT can select
a significantly lower dimensional feature subspace distinctly and
robustly while maintaining high decision performance. 

The rest of this paper is organized as follows. Related previous work is
reviewed in Sec.  \ref{sec:review}. DFT and RFT are presented in Sec.
\ref{sec:method}.  Experimental results are shown in Sec.
\ref{sec:experiments}. Finally, concluding remarks are given in Sec.
\ref{sec:conclusion}. 

\section{Review of Previous Work}\label{sec:review}

Feature selection methods can be categorized into unsupervised
\cite{mitra2002unsupervised, cai2010unsupervised, qian2013robust,
solorio2020review}, semi-supervised \cite{sheikhpour2017survey,
zhao2007semi}, and supervised \cite{huang2015supervised} three types.
Unsupervised methods focus on the statistics of input features while
ignoring the target class or value. Straightforward unsupervised methods
can be fast, e.g., removing redundant features using correlation,
removing features of low variance.  However, their power is limited and
less effective than supervised methods. More advanced unsupervised
methods adopt clustering. Examples include \cite{lee1999learning,
hartigan1972direct, aggarwal1999fast}.  Their complexity is higher,
their behavior is not well understood, and their performance is not easy
to evaluate systematically. Overall, this is an open research field. 

Existing semi-supervised and supervised feature selection methods can be
classified into wrapper, filter and embedded three classes
\cite{sheikhpour2017survey}. Wrapper methods \cite{kohavi1997wrappers}
create multiple models with different subsets of input features and
select the model containing the features that yield the best
performance. One example is recursive feature elimination (RFE)
\cite{rfe}. This process can be computationally expensive.  Filter
methods involve evaluating the relationship between input and target
variables using statistics and selecting those variables that have the
strongest relation with the target ones.  One example is the analysis of
variance (ANOVA) \cite{anova}. This approach is computationally
efficient with robust performance.  Another example is feature
selection based on linear discriminant analysis (LDA). It finds the most
separable projection directions. The objective function of LDA is used
to select discriminant features from the existing feature dimensions by
measuring the ratio between the between-class scatter matrix and the
within-class scatter matrix. It can be generalized from the 2-class
problem to the multi-class problem. Embedded methods perform feature
selection in the process of training and are usually specific to a
single learner. One example is ``feature importance" (FI) obtained from
the training process of the XGBoost classifier/regressor \cite{xgb},
which is also known as ``feature selection from model" (SFM). 

Inspired by information theory and the decision tree, a novel
supervised feature selection methodology is proposed in this work. The
resulting tests are called the discriminant feature test (DFT) and the
relevant feature test (RFT) for classification and regression tasks,
respectively. Our proposed methods belong to the filter methods, which
give a score to each dimension and select features based on feature
ranking. The scores are measured by the weighted entropy and the
weighted MSE for DFT and RFT, which reflect the discriminant power and
relevance degree to classification and regression targets,
respectively.

To demonstrate the power of DFT and RFT, we conduct performance
benchmarking between DFT/RFT, ANOVA and FI from XGBoost in the
experimental section.  To this end, we use deep features obtained by
LeNet-5 for MNIST and Fashion-MNIST datasets as illustrative examples.
Other datasets with handcrafted features and gene expressions features
are also used for performance benchmarking. Comparison with the
minimal-redundancy-maximal-relevance (mRMR)
criterion~\cite{peng2005feature, ding2005minimum}, which is a more
advanced feature selection method, is also conducted.  It is shown by
experimental results that DFT and RFT can select a lower dimensional
feature subspace distinctly and robustly while maintaining high decision
performance.

%%%%%%%%%%%%%%%%%%%%%%%%%%%%%%%%%%%%%%%%%%%%%%%%%%%%%%%%%%%%%%%%%%
\begin{figure*}[tb]
\centerline{\includegraphics[width=0.95\linewidth]{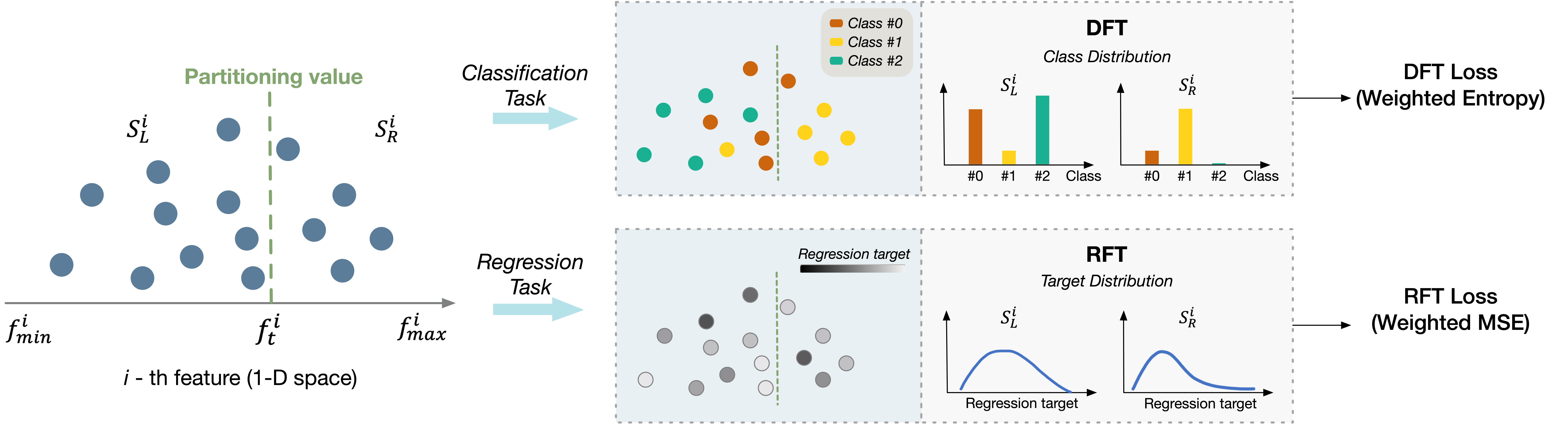}}
\caption{An overview of the proposed feature selection methods: DFT and
RFT. For the $i$-th feature, DFT measures the class distribution in
$S^i_L$ and $S^i_R$ to compute the weighted entropy as the DFT loss,
while RFT measures the weighted estimated regression MSE in both sets as
the RFT loss.}
% \vspace{-5pt}
\label{fig:pipeline}
\end{figure*}
%%%%%%%%%%%%%%%%%%%%%%%%%%%%%%%%%%%%%%%%%%%%%%%%%%%%%%%%%%%%%%%%%%

\section{Proposed Feature Selection Methods}\label{sec:method}

Being motivated by the feature selection process in the decision tree
classifier, we propose two feature selection methods, DFT and RFT, in
this section as illustrated in Fig. \ref{fig:pipeline}. They will be
detailed in Sec. \ref{subsec:DFT} and Sec. \ref{subsec:RFT},
respectively. Finally, robustness of DFT and RFT will be discussed
in Sec. \ref{sec:robustness}.

\subsection{Discriminant Feature Test (DFT)} \label{subsec:DFT}

Consider a classification problem with $N$ data samples, $P$ features
and $C$ classes. Let $f^i$, $1 \le i \le P$, be a feature dimension and
its minimum and maximum are $f_{\min}^i$ and $f_{\max}^{i}$,
respectively.  DFT is used to measure the discriminant power of each
feature dimension out of a $P$-dimensional feature space independently.
If feature $f^i$ is a discriminant one, we expect data samples projected
to it should be classified more easily. To check it, one idea is to
partition $[f_{\min}^i, f_{\max}^{i}]$ into $M$ nonoverlapping
subintervals and adopt the maximum likelihood rule to assign the class
label to samples inside each subinterval. Then, we can compute the
percentage of correct predictions. The higher the prediction accuracy,
the higher the discriminant power. Although prediction accuracy may
serve as an indicator for purity, it does not tell the distribution of
the remaining $C-1$ classes if $C>2$. Thus, it is desired to consider
other purity measures. 

In our design, we use the weighted entropy of the left and right
subsets as the DFT loss to measure the discriminant power of each
dimension. The reason of choosing the weighted entropy as the cost is
that it considers the probability distribution of all classes instead of
the maximum likelihood rule in prediction accuracy as stated above. A
lower entropy value is obtained from a more biased distribution of
classes, indicating the subinterval is dominated by fewer classes.

By following the practice of a binary decision tree, we consider the
case, $M=2$, as shown in the left subfigure of Fig. \ref{fig:pipeline},
where $f_t^{i}$ denotes the threshold position of two sub-intervals.  If
a sample with its $i$th dimension, $x^{i}_n < f_t^{i}$, it goes to the
subset associated with the left subinterval. Otherwise, it will go to
the subset associated with the right subinterval. Formally, the
procedure of DFT consists of three steps for each dimension as detailed
below. 

\subsubsection{Training Sample Partitioning}
For the $i$th feature, $f^{i}$, we need to search for the optimal
threshold, $f^{i}_{op}$, between $[f_{\min}^i, f_{\max}^{i}]$ and
partition training samples into two subsets $S^{i}_L$ and $S^{i}_R$ via
\begin{eqnarray}\label{eq:eq1}
&& \mbox{if} \, x^{i}_n < f^{i}_{op}, \; x_n\, \epsilon \, S^{i}_L; \; \;  \\
&& \mbox{otherwise}, \; x_n\, \epsilon \, S^{i}_R,
\end{eqnarray}
where $x^{i}_n$ represents the $i$-th feature of the $n$-th training sample $x_n$, and $f^{i}_{op}$ is selected automatically to optimize a certain
purity measure.  To limit the search space of $f^{i}_{op}$, we partition
the entire feature range, $[f_{\min}^i, f_{\max}^{i}]$, into $B$ uniform
segments and search the optimal threshold among the following $B-1$
candidates:
\begin{equation}\label{eq:threshold_candidate}
f_{b}^i=f_{\min}^i+ \frac{b}{B} [f_{\max}^{i}-f_{\min}^i], \quad 
b=1,\cdots, B-1,
\end{equation}
where $B=2^j$, $j=1,2 \cdots$, is examined in Sec.~\ref{sec:robustness}.

\subsubsection{DFT Loss Measured by Entropy} 
Samples of different classes belong to $S^{i}_L$ or $S^{i}_R$. Without loss of generality, the following discussion is based on the assumption that each class has the same number of samples in the full training set; namely $S^{i}_L \cup S^{i}_R$. To
measure the purity of subset $S^{i}_L$ corresponding to the partition
point $f^{i}_{t}$, we use the following entropy metric:
\begin{equation}\label{eq:entropy}
H^{i}_{L,t} = -\sum_{c=1}^{C}p^{i}_{L,c}log(p^{i}_{L,c}),
\end{equation}
where $p^{i}_{L,c}$ is the probability of class $c$ in $S^{i}_L$.
Similarly, we can compute entropy $H^{i}_{R,t}$ for subset $S^{i}_R$.
Then, the entropy of the full training set against
partition $f^{i}_t$ is the weighted average of $H_{L,t}$ and $H_{R,t}$
in form of
\begin{equation}\label{eq:weighted_entropy}
H^{i}_t = \frac{N^{i}_{L,t} H^{i}_{L,t}+ N^{i}_{R,t} H^{i}_{R,t}}{N},
\end{equation}
where $N^{i}_{L,t}$ and $N^{i}_{R,t}$ are the sample numbers in subsets
$S^{i}_L$ and $S^{i}_R$, respectively, and $N=N^{i}_{L,t}+N^{i}_{R,t}$
is the total number of training samples. The optimized entropy $H^i_{op}$
for the $i$-th feature is given by
\begin{equation}\label{eq:optimized_entropy}
H^i_{op} = \min_{t \epsilon T} H^i_{t},
\end{equation}
where $T$ indicates the set of partition points.

\subsubsection{Feature Selection Based on Optimized Loss}
We conduct search for optimized entropy values, $H^i_{op}$, of all
feature dimensions, $f^i$, $1 \le i \le P$ and order the values of
$H^i_{op}$ from the smallest to the largest ones. The lower the
$H^i_{op}$ value, the more discriminant the $i$th-dimensional feature,
$f^i$. Then, we select the top $K$ features with the lowest entropy
values as discriminant features. To choose the value of $K$ with little
ambiguity, it is critical the rank-ordered curve of $H^i_{op}$ should
satisfy one important criterion. That is, it should have a distinct and
narrow elbow region. We will show that this is indeed the case in
Sec. \ref{sec:experiments}.

\subsection{Relevant Feature Test (RFT)}\label{subsec:RFT}

For regression tasks, the mapping between an input feature and a target
scalar function can be more efficiently built if the feature dimension
has the ability to separate samples into segments with smaller
variances.  This is because the regressor can use the mean of each
segment as the target value, and its corresponding variance indicates
the prediction mean squared-error (MSE) of the segment. Motivated by
this observation and the binary decision tree, RFT partitions a feature
dimension into the left and right two subintervals and evaluates the
total MSE from them. We use this approximation error as the RFT loss
function.  The smaller the RFT loss, the better the feature dimension.
Again, the RFT loss depends on the threshold $f_t^{i}$. The process of
selecting more powerful feature dimensions for regression is named
Relevant Feature Test (RFT).  Similar to DFT, RFT has three steps. They
are elaborated below.  Here, we adopt the same notations as those in
Sec.  \ref{subsec:DFT}. 

\subsubsection{Training Sample Partitioning} 

By following the first step in DFT, we search for the optimal threshold,
$f^{i}_{op}$, between $[f_{\min}^i, f_{\max}^{i}]$ and partition
training samples into two subsets $S^{i}_L$ and $S^{i}_R$ for the $i$th
feature, $f^{i}$. Again, we partition the feature range, $[f_{\min}^i,
f_{\max}^{i}]$, into $B$ uniform segments and search the optimal
threshold among the following $B-1$ candidates as given in Eq.
(\ref{eq:threshold_candidate}). 

\subsubsection{RFT Loss Measured by Estimated Regression MSE} 

We use $y$ to denote the regression target value. For the $i$th feature
dimension, $f^{i}$, we partition the sample space into two disjoint ones
$S^{i}_L$ and $S^{i}_R$. Let $y^i_L$ and $y^i_R$ be the mean of target
values in $S^{i}_L$ and $S^{i}_R$, and we use $y^i_L$ and $y^i_R$ as the estimated
regression value of all samples in $S^{i}_L$ and $S^{i}_R$,
respectively. Then, the RFT loss is defined as the sum of estimated regression
MSEs of $S^{i}_L$ and $S^{i}_R$. Mathematically, we have
\begin{equation}\label{eq:weighted_MSE}
R^{i}_t = \frac{N^{i}_{L,t} R^{i}_{L,t} + N^{i}_{R,t} R^{i}_{R,t}}{N},
\end{equation}
where $N=N^{i}_{L,t}+N^{i}_{R,t}$, $N^{i}_{L,t}$, $N^{i}_{R,t}$,
$R^{i}_{L,t}$ and $R^{i}_{R,t}$ denote the sample numbers and the estimated regression
MSEs in subsets $S^{i}_L$ and $S^{i}_R$, respectively. Feature $f^{i}$
is characterized by its optimized estimated regression MSE over the set, $T$, of
candidate partition points:
\begin{equation}\label{eq:optimized_MSE}
R^i_{op} = \min_{t \epsilon T} R^i_{t}.
\end{equation}

\subsubsection{Feature Selection Based on Optimized Loss}

We order the optimized estimated regression MSE value, $R^i_{op}$ across all
feature dimensions, $f^i$, $1 \le i \le P$, from the smallest to the
largest ones. The lower the $R^i_{op}$ value, the more relevant the
$i$th-dimensional feature, $f^i$.  Afterwards, we select the top $K$
features with the lowest estimated regression MSE values as relevant features. 

\subsection{Robustness Against Bin Numbers}\label{sec:robustness}

\begin{figure}[tbp]
\centerline{\includegraphics[width=0.65\linewidth]{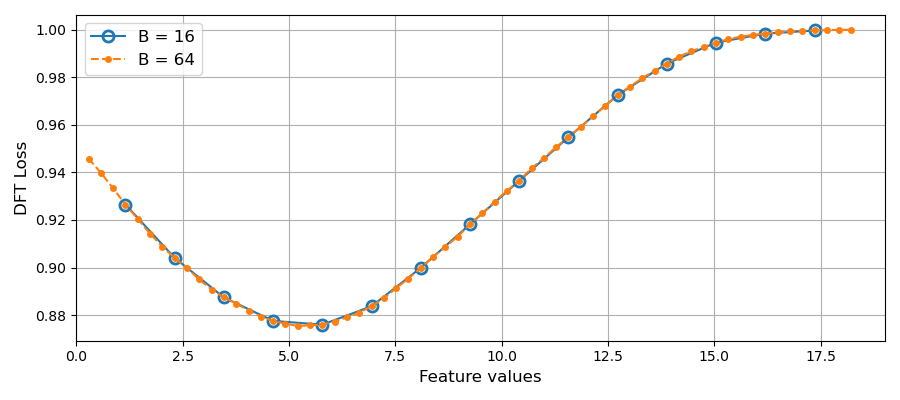}}%\\
\vspace{-5pt}
\centerline{(a)} %\\
\centerline{\includegraphics[width=0.65\linewidth]{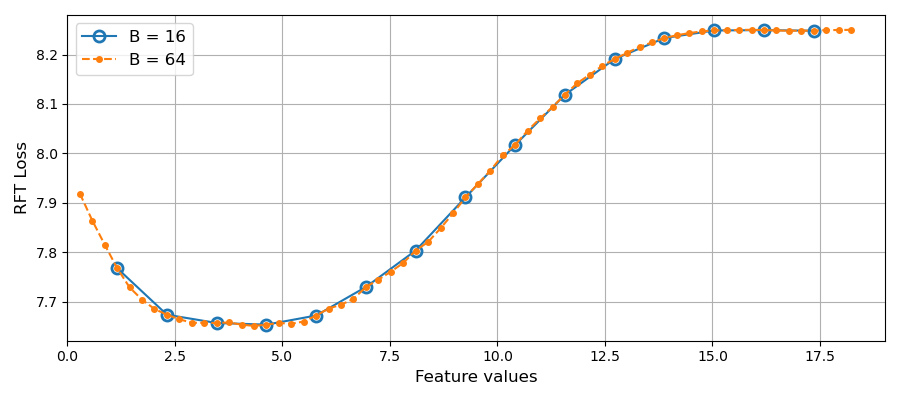}}%\\
\vspace{-5pt}
\centerline{(b)} %\\
% \vspace{-5pt}
\caption{Comparison of two binning schemes with $B=16$ and $B=64$: (a) DFT and (b) RFT.}
\label{fig:embedding}
\end{figure}

For smooth DFT/RFT loss curves with a sufficiently large bin number
(say, $B \ge 16$), the optimized loss value does not vary much by
increasing $B$ furthermore as shown in Fig.~\ref{fig:embedding}.
Figs.~\ref{fig:embedding}(a) and (b) show the DFT and RFT loss functions
for an exemplary feature, $f^i$, under two binning schemes; i.e.,
$B=16$ and $B=64$, respectively.  We see that the binning $B=16$ is fine
enough to locate the optimal partition $f^{i}_{op}$.  If $B=2^j$,
$j=1,2, \cdots$, the set of partition points in a small $B$ value is a
subset of those of a larger $B$ value.  Generally, we have the following
observations. The difference of the DFT/RFT loss between adjacent
candidate points changes smoothly. Since the global minimum has a flat
bottom, the loss function is low for a range of partition thresholds.  The
feature will achieve a similar loss level with multiple binning schemes.
For example, Fig.~\ref{fig:embedding}(a) shows that $B=16$ reaches the
global minimum at $f^i=5.21$ while $B=64$ reaches the global minimum at
$f^i=5.78$. The difference is about 3\% of the full dynamic range of
$f^i$. Similar characteristics are observed for all feature dimensions
in DFT/RFT, indicating the robustness of DFT/RFT. For lower
computational complexity, we typically choose $B=16$ or $B=32$. 

% %%%%%%%%%%%%%%%%%%%%%%%%%%%%%%%%%%%%%%%%%%
%%%%%%%%%%%%%%%%%%%%%%%%%%%%%%%%%%%%%%%%%%%

\section{Experimental Results}\label{sec:experiments}

\subsection{Image Datasets with High Dimensional Feature Space}

To demonstrate the power of DFT and RFT, we consider several
classical datasets. They include MNIST~\cite{lecun1998gradient},
Fashion-MNIST~\cite{xiao2017fashion}, the Multiple Features (MultiFeat)
dataset, the Arrhythmia (ARR) dataset~\cite{guvenir1997supervised} from
the UCI machine learning archive~\cite{Dua:2019}, and the Colon cancer
dataset~\cite{alon1999broad}. The latter three are used to measure DFT
in the classification problem setting.

\textbf{Dataset-1: MNIST and Fashion-MNIST.} Both datasets contain
grayscale images of resolution $28\times28$, with 60K training and 10K
test images.  MNIST has 10 classes of hand-written digits (from 0 to 9)
while Fashion-MNIST has 10 classes of fashion products. In order to get
deep features for each dataset, we train the LeNet-5
network~\cite{lecun1998gradient} for the two corresponding
classification problems and adopt the 400-D feature vector before the
two FC layers as raw features to evaluate several feature selection
methods.  Besides original clean training images, we add additive
zero-mean Gaussian noise with different standard deviation values to
evaluate the robustness of feature selection methods against noisy data.
The LeNet-5 network is re-trained for these noisy images and the
corresponding deep features are extracted. For the performance
benchmarking purpose, we list the test classification accuracy of the
trained LeNet-5 for MNIST and Fashion-MNIST in Table~\ref{tab:MLP_acc}
to illustrate the quality of the deep features. 
%%%%%%%%%%%%%%%%%%%%%%%%%%%%%%%%%%%%%%%%%%%%%%%%%%%%%%%%%%%%%%%%%%%%
\begin{table}%[t]
\begin{center}
\caption{Classification accuracy (\%) of LeNet-5 on MNIST and
Fashion-MNIST.\label{tab:MLP_acc}}
{\fontsize{10}{12}\selectfont\begin{tabular}{lcc}
\toprule
              & Clean & Noisy \\
\midrule
MNIST         & 99.18 & 98.85 \\
Fashion-MNIST & 90.19 & 86.95 \\ 
\bottomrule
\end{tabular}}
\end{center}
\end{table}
%%%%%%%%%%%%%%%%%%%%%%%%%%%%%%%%%%%%%%%%%%%%%%%%%%%%%%%%%%%%%%%%%%%%

\textbf{Dataset-2: MultiFeat}~\cite{van1998handwritten,
van1998neural,jain1997feature}. This dataset contains features of
hand-written digits (from 0 to 9) extracted from a collection of Dutch
utility maps ~\cite{Dua:2019}, including 649 dimensional features for
200 images per class. Different from the deep features in Dataset-1, the
649 features are extracted from six perspectives such as Fourier
coefficients of character shapes and morphological features. Since the
number of samples is small, we use 10-fold cross-validation and compute
the mean accuracy to evaluate the classification performance.

\textbf{Dataset-3: Colon.} This gene expression dataset contains 62
samples with 2000 features each. It has a binary classification label;
namely, the normal tissue or the cancerous tissue. There are 22 normal
tissue and 40 cancer tissue samples. Considering its small sample size,
we use the leave-one-out validation to get the classification
predictions for each sample.

\textbf{Dataset-4: ARR.} This cardiac arrhythmia dataset has binary
labels for 237 normal and 183 abnormal samples. Each sample contains 278
features. The 10-fold cross-validation is adopted to evaluate the
classification performance.

\subsection{DFT for Classification Problems}

We compare effectiveness of four feature selection methods: 1) F scores
from ANOVA (ANOVA F Scores), 2) absolute correlation coefficient w.r.t
the class labels (Abs. Corr. Coeff.), 3) feature importance (Feat. Imp.)
from a pre-trained XGBoost classifier, and 4) DFT.  We adopt four
classifiers to validate the classification performance. They are the
Logistic Regression (LR) classifier~\cite{logisticregression}, the
Support Vector Machine (SVM) classifier~\cite{svm}, the Random Forest
(RF) classifier~\cite{RF}, and the XGBoost classifier~\cite{xgb}. We
have the following two observations. 

\subsubsection{DFT offers an obvious elbow point}
\begin{figure*}[tbp]
\centerline{\includegraphics[width=0.65\linewidth]{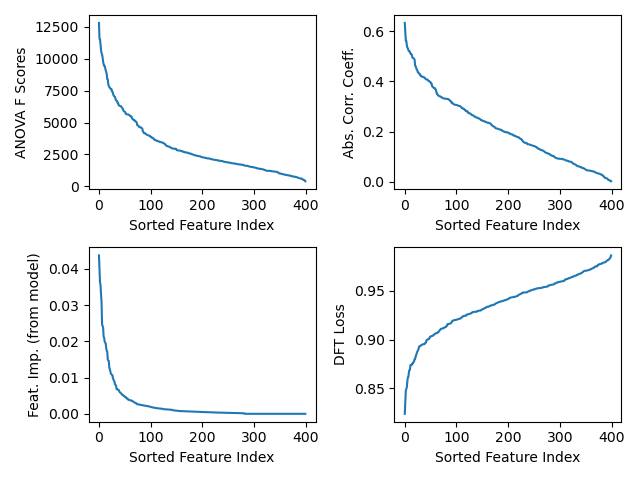}}
\vspace{-5pt}
\caption{Comparison of distinct feature selection capability among
four feature selection methods for the classification task on the 
Fashion-MNIST dataset.}
\vspace{-10pt}
\label{fig:clf_fashion_4_featimp}
\end{figure*}

Fig.~\ref{fig:clf_fashion_4_featimp} compares the ranked scores of four
feature selection methods on Fashion-MNIST dataset. The lower DFT loss,
the higher importance of a feature. The other three have a reversed
relation, namely, the higher the score, the higher the importance. Thus,
we search for the elbow point for DFT but the knee point for the other
methods. Clearly, the feature importance curve from the pre-trained
XGBoost classifier has a clearer knee point and the DFT curve has a more
obvious elbow point. In contrast, ANOVA and
correlation-coefficient-based methods are not as effective in selecting
discriminant features since their knee points are less obvious.

% %%%%%%%%%%%%%%%%%%%%%%%%%%%%%%%%%%%%%%%%%%%%%%%%%%%%%%%%%%%%%%%%%%%%
% %%%%%%%%%%%%%%%%%%%%%%%%%%%%%%%%%%%%%%%%%%%%%%%%%%%%%%%%%%%%%%%%%%%%

\subsubsection{Features selected by DFT achieves comparable and stable
classification performance}

Tables~\ref{tab:DFT_performance_mnist_clean} -
\ref{tab:DFT_performance_fashion_noisy} summarize the classification
accuracy using four classifiers at two reduced dimensions selected by
the DFT loss curve based on early and late elbow points on
Dataset-1. The RBF kernel is used for SVM. We see that DFT can achieve
comparable (or even the best) performance among the four methods at the
same selected feature dimension. The accuracy gap between the late
elbow point and the full feature set (400-D) is very small.  They are
0.62\% and 0.94\% using XGBoost classifier for clean MNIST and
Fashion-MNIST, respectively. The late elbow point only uses 25-35\% of
the full feature set. The gaps in classification accuracy on noisy
images are 0.58\% and 1.6\% for MNIST and Fashion-MNIST, respectively,
indicating the robustness of the DFT feature selection method against
input perturbation. 

We also show the classification performance on the Colon dataset using LR
and SVM in Table~\ref{tab:colon_performance}, where the linear kernel is
used in SVM. DFT has the minimum or a comparable number of errors in
leave-one-out validation. Furthermore, DFT can always achieve fewer
errors as compared to the setting of using all 2000 features. 

% %%%%%%%%%%%%%%%%%%%%%%%%%%%%%%%%%%%%%%%%%%%%%%%%%%%%%%%%%%%%%%%%%%%%
\begin{table*}[tbp]
	\centering
%	\begin{center}
		\caption{Comparison of classification performance (\%) on \textbf{Clean} 
MNIST between different feature selection methods.}\label{tab:DFT_performance_mnist_clean}
		{\fontsize{10}{12}\selectfont\begin{tabular}{@{}clcccc@{}}
			\toprule
		Selected Dimension&
		Method&
		LR &
		SVM &
		RF &
		XGBoost \\
		\midrule
		&ANOVA &
		94.21 &
		95.07 &
		95.77 &
		96.58 \\

		Early Elbow Point&
		Corr. &
		88.73 &
		92.47 &
		94.04 &
		95.11 \\

		(30-D)&
		Feat. imp. &
		92.61 &
		93.55 &
		94.89 &
		95.71 \\

		&DFT (Ours) &
		\textbf{94.49} &
		\textbf{95.45} &
		\textbf{96.29} &
		\textbf{96.92} \\[6pt]
		&ANOVA &
  		98.24 &
		\textbf{98.22} &
		97.98 &
		98.66 \\

		Late Elbow Point&Corr. &
		97.61 &
		97.78 &
		97.35 &
		98.57 \\

		(100-D)&
		Feat. imp. &
		\textbf{98.24} &
		98.15 &
		\textbf{98.18} &
		\textbf{98.78} \\
		
		&DFT (Ours) &
		97.93 &
		97.83 &
		97.81 &
		98.52 \\[6pt]
		Full Set (400-D)&
		&98.89 &
		98.77&
		98.61&
		99.14\\ 
	\bottomrule
	\end{tabular}}
%\end{center}
\end{table*}

% %%%%%%%%%%%%%%%%%%%%%%%%%%%%%%%%%%%%%%%%%%%%%%%%%%%%%%%%%%%%%%%%%%%%

% %%%%%%%%%%%%%%%%%%%%%%%%%%%%%%%%%%%%%%%%%%%%%%%%%%%%%%%%%%%%%%%%%%%%
\begin{table*}[tbp]
	\centering
	%	\begin{center}
	\caption{Comparison of classification performance (\%) on \textbf{Noisy} MNIST between 
different feature selection methods.}\label{tab:DFT_performance_mnist_noisy}
	{\fontsize{10}{12}\selectfont\begin{tabular}{@{}clcccc@{}}
			\toprule
			Selected Dimension&
			Method&
			LR &
			SVM &
			RF &
			XGBoost \\
			\midrule
			&ANOVA &
			\textbf{94.22} &
			94.62 &
			95.60 &
			96.04 \\
			
			Early Elbow Point&
			Corr. &
			90.97 &
			93.06 &
			93.64 &
			95.21 \\
			
			(40-D)&
			Feat. imp. &
  			92.59 &
			93.35 &
			94.48 &
			95.34 \\
			
			&DFT (Ours) &
  			\underline {94.03} &
			\textbf{95.22} &
			\textbf{95.78} &
			\textbf{96.59} \\[6pt]
			&ANOVA &
  			96.81 &
			96.87 &
			97.16 &
			97.99 \\
			
			Late Elbow Point&Corr. &
  			96.87 &
			97.13 &
			96.83 &
			97.93 \\
			
			(100-D)&
			Feat. imp. &
  			\textbf{97.22} &
			97.2 &
			97.36 &
			97.97 \\
			
			&DFT (Ours) &
  			\underline {97.08} &
			\textbf{97.36} &
			\textbf{97.49} &
			\textbf{98.18} \\[6pt]
			Full Set (400-D)&
			&98.04 &
			98.17 &
			98.15 &
			98.76 \\
			\bottomrule
	\end{tabular}}
	%\end{center}
\end{table*}

% %%%%%%%%%%%%%%%%%%%%%%%%%%%%%%%%%%%%%%%%%%%%%%%%%%%%%%%%%%%%%%%%%%%%

% %%%%%%%%%%%%%%%%%%%%%%%%%%%%%%%%%%%%%%%%%%%%%%%%%%%%%%%%%%%%%%%%%%%%
\begin{table*}[tbp]
	\centering
	
	%	\begin{center}
	\caption{Comparison of classification performance (\%) on \textbf{Clean} Fashion-MNIST 
between different feature selection methods.}\label{tab:DFT_performance_fashion_clean}
	{\fontsize{10}{12}\selectfont\begin{tabular}{@{}clcccc@{}}
			\toprule
			Selected Dimension&
			Method&
			LR &
			SVM &
			RF &
			XGBoost \\
			\midrule
			&ANOVA &
			78.85 &
			80.44 &
			83.33 &
			83.11 \\
			
			Early Elbow Point&
			Corr. &
			76.57 &
			80.16 &
			82.69 &
			83.04 \\
			
			(30-D)&
			Feat. imp. &
			78.96 &
			80.49 &
			82.99 &
			82.96 \\
			
			&DFT (Ours) &
			\textbf{79.59} &
			\textbf{81.48} &
			\textbf{84.03} &
			\textbf{84.09} \\
			
			\\ 
			
			&ANOVA &
			 87.06 &
			86.61 &
			87.69 &
			89.08  \\
			
			Late Elbow Point&Corr. &
			86.99 &
			86.96 &
			87.36 &
			88.81 \\
			
			(150-D)&
			Feat. imp. &
			87.47 &
			\textbf{87.62} &
			\textbf{88.28} &
			\textbf{89.33} \\
			
			&DFT (Ours) &
			\textbf{87.60} &
			\underline{87.02} &
			\underline{87.71} &
			\underline{89.13} \\
			\\
			Full Set (400-D)&
			&89.05 &
			88.18 &
			88.74 &
			90.07\\ 
			\bottomrule
	\end{tabular}}
	%\end{center}
\end{table*}

% %%%%%%%%%%%%%%%%%%%%%%%%%%%%%%%%%%%%%%%%%%%%%%%%%%%%%%%%%%%%%%%%%%%%

% %%%%%%%%%%%%%%%%%%%%%%%%%%%%%%%%%%%%%%%%%%%%%%%%%%%%%%%%%%%%%%%%%%%%
\begin{table*}[htbp]
	\centering
	\small
	%	\begin{center}
	\caption{Comparison of classification performance (\%) on \textbf{Noisy} Fashion-MNIST 
between different feature selection methods.}\label{tab:DFT_performance_fashion_noisy}
	{\fontsize{10}{12}\selectfont\begin{tabular}{@{}clcccc@{}}
			\toprule
			Selected Dimension&
			Method&
			LR &
			SVM &
			RF &
			XGBoost \\
			\midrule
			&ANOVA &
			75.35 &
			76.41 &
			77.94 &
			78.62 \\
			
			Early Elbow Point&
			Corr. &
			75.55 &
			\textbf{77.94} &
			79.22 &
			\textbf{80.50}\\
			
			(40-D)&
			Feat. imp. &
			75.73 &
			77.1 &
			78.06 &
			78.63 \\
			
			&DFT (Ours) &
			\textbf{76.35} &
			\underline{77.92} &
			\textbf{79.23} &
			\underline{79.69} \\
			
			\\ 
			
			&ANOVA &
			81.84 &
			81.98 &
			82.42 &
			84.10 \\
			
			Late Elbow Point&Corr. &
			82.26 &
			82.9 &
			82.59 &
			84.72  \\
			
			(150-D)&
			Feat. imp. &
			\textbf{83.19} &
			\textbf{83.43} &
			\textbf{83.54} &
			\textbf{84.91} \\
			
			&DFT (Ours) &
			 82.08 &
			82.40 &
			\underline{82.61} &
			84.31 \\
			\\
			Full Set (400-D)&
			&84.35 &
			84.24 &
			84.23 &
			85.91 \\
			\bottomrule
	\end{tabular}}
	%\end{center}
\end{table*}

\begin{table*}[tbp]
    \small
	
	\centering
	\caption{Comparison of number of errors on Colon cancer dataset between 
different feature selection methods.}
	\label{tab:colon_performance}
	{\fontsize{10}{12}\selectfont\begin{tabular}{@{}llccccccc@{}}
		\toprule	
		\multirow{2}{*}{Classifier}&\multirow{2}{*}{Method}         & \multicolumn{6}{c}{Selected Dimension} & \multicolumn{1}{l}{Full Set} \\
		&&5&10&20&50&80&100&2000-D\\
		\midrule
		\multirow{4}{*}{LR}  & ANOVA      & 6          & 9          & 10         & 10         & 7          & 7          & \multirow{4}{*}{10} \\
		& Corr.      & 6          & 9          & 10         & 10         & 7          & 7          &                     \\
		& Feat. Imp. & 8          & \textbf{6} & \textbf{5} & \textbf{5} & 9          & 9          &                     \\
		& DFT (Ours) & \textbf{6} & \underline{7}    & \underline{9}    & \underline{8}    & \textbf{7} & \textbf{7} &                     \\ \\
		
		\multirow{4}{*}{SVM} & ANOVA      & 6          & 7          & 7          & 9          & 8          & 9          & \multirow{4}{*}{9}  \\
		& Corr.      & 6          & 7          & 7          & 9          & 8          & 9          &                     \\
		& Feat. Imp. & 6          & 6          &  \textbf{5}        & \textbf{6} & 9          & 8          &                     \\
		& DFT (Ours) & \textbf{6} & \textbf{6} & 8  & \underline{8}    & \textbf{6} & \textbf{6} &                     \\ 
		\bottomrule
	\end{tabular}}
\end{table*}
% %%%%%%%%%%%%%%%%%%%%%%%%%%%%%%%%%%%%%%%%%%%%%%%%%%%%%%%%%%%%%%%%%%%%

\subsubsection{Comparison between DFT and mRMR}

The minimal-redundancy-maximal-relevance
(mRMR) \cite{peng2005feature,ding2005minimum} aims at finding a feature
set with high relevance to the class while keeping the selected features
with small redundancy, leading to an efficient but effective subset of
features. It combines constraints measured by the mutual information of
both relevance to the class and redundancy between selected features and
treats it as an optimization problem. In this experiment, we compare DFT
with mRMR using its incremental selection scheme.

\begin{figure*}[tbp]
\centering
\begin{subfigure}{0.45\textwidth}
\centering
    \includegraphics[width=0.95\linewidth]{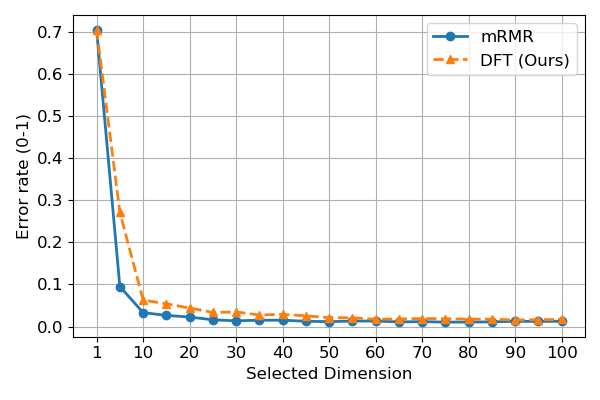}
    \caption{SVM}
\end{subfigure}
\begin{subfigure}{0.45\textwidth}
\centering
    \includegraphics[width=0.95\linewidth]{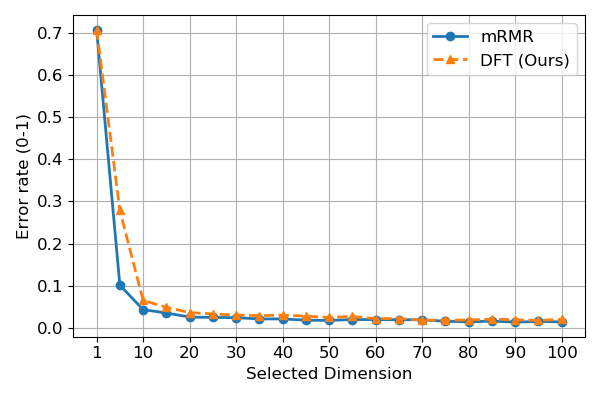}
    \caption{XGBoost}
\end{subfigure}
\vspace{-5pt}
\caption{Error rate comparison on the MultiFeat dataset between mRMR and DFT.}
\label{fig:err_comparison_mRMR_multiFeat}
\end{figure*}

\begin{figure*}[tbp]
\centering
\begin{subfigure}{0.45\textwidth}
\centering
    \includegraphics[width=0.95\linewidth]{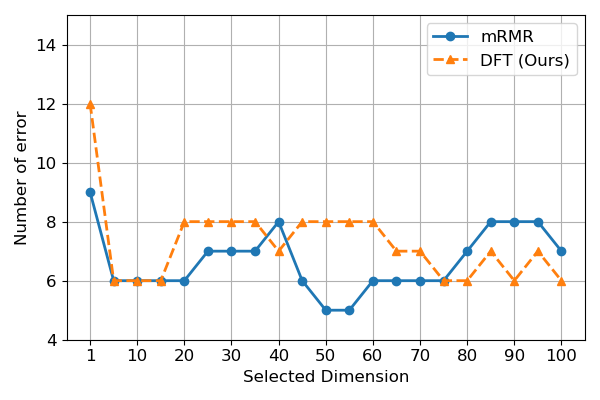}
    \caption{SVM}
\end{subfigure}
\begin{subfigure}{0.45\textwidth}
\centering
    \includegraphics[width=0.95\linewidth]{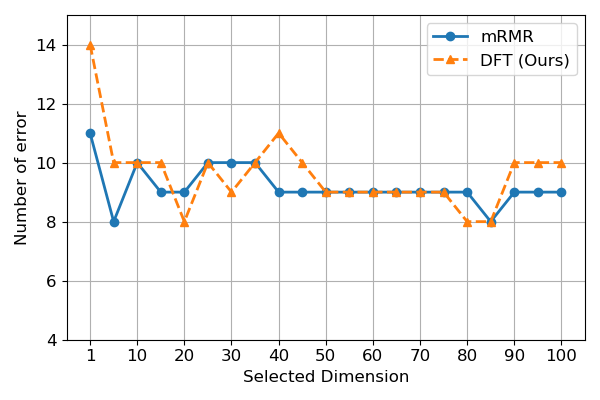}
    \caption{XGBoost}
\end{subfigure}
\vspace{-5pt}
\caption{Comparison of the number of errors on the Colon dataset between mRMR and DFT.}
\label{fig:err_comparison_mRMR_colon}
\end{figure*}

\begin{figure*}[!htbp]
\centering
\begin{subfigure}{0.45\textwidth}
\centering
    \includegraphics[width=0.95\linewidth]{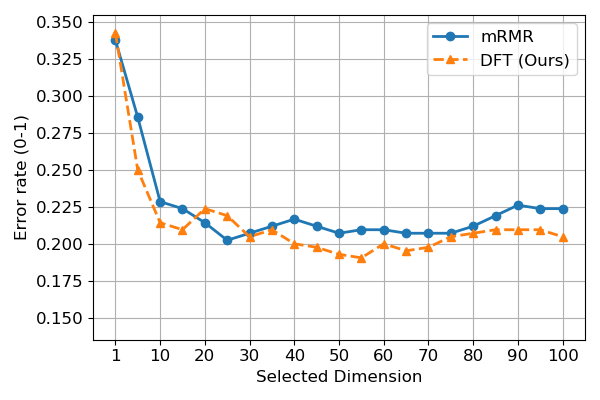}
    \caption{SVM}
\end{subfigure}
\begin{subfigure}{0.45\textwidth}
\centering
    \includegraphics[width=0.95\linewidth]{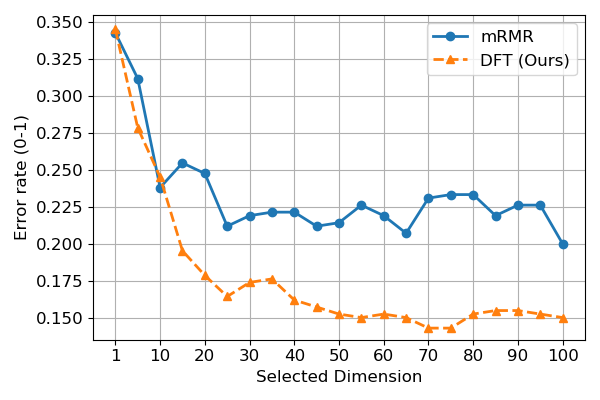}
    \caption{XGBoost}
\end{subfigure}
\vspace{-5pt}
\caption{Error rate comparison on the ARR dataset between mRMR and the DFT.}
\label{fig:err_comparison_mRMR_arr}
\end{figure*}

Figs.~\ref{fig:err_comparison_mRMR_multiFeat}-\ref{fig:err_comparison_mRMR_arr}
compare the performance of mRMR and DFT on MultiFeat, Colon and
ARR datasets with SVM and XGBoost classifiers. For MultiFeat and
Colon, DFT can achieve very competitive performance with mRMR.
Specifically, the most discriminant feature of MultiFeat selected by DFT
is identical to the first feature selected by mRMR. The error rate with
the top 5 features selected by mRMR is smaller than that of DFT. Yet, the
performance gap is substantially narrowed after selecting more than 10
features out of the total 649 features.  Overall,
the error rate of DFT and mRMR converges at similar reduced dimensions, as shown in Figs.~\ref{fig:err_comparison_mRMR_multiFeat} and
\ref{fig:err_comparison_mRMR_arr}.  On the other hand, the
error rate of DFT on the ARR dataset is much lower than that of mRMR,
with around 2.5\% and 5\% gap at 100 selected features for SVM and
XGBoost, respectively, as shown in Fig. \ref{fig:err_comparison_mRMR_arr}.

\subsubsection{DFT requires less running time}

We compare time efficiency of DFT, ANOVA, mRMR and feature
importance from the XGBoost model. Table~\ref{tab:running_time}
summarizes the running time for MultiFeat, Colon and ARR datasets. All
methods are run on the same CPU.  The pre-trained XGBoost classifier
uses the maximum depth of one with 300 trees. For filter methods such as
ANOVA and DFT, the time is evaluated on all features without parallel
computing.  For mRMR, we set the maximum to 100 for incremental
selection, which is smaller than the full feature set. ANOVA is the
fastest and DFT method ranks the second on all three datasets.  The
running time of DFT with $B=16$ is about $\times2.6$, $\times7.3$ and
$\times23.1$ times faster than mRMR on MultiFeat, Colon and ARR,
respectively. To further reduce the running time, our proposed DFT can
be easily improved by adopting parallel computing since it processes
each feature independently before the feature ranking.

% %%%%%%%%%%%%%%%%%%%%%%%%%%%%%%%%%%%%%%%%%%%%%%%%%%%%%%%%%%%%%%%%%%%%
% %%%%%%%%%%%%%%%%%%%%%%%%%%%%%%%%%%%%%%%%%%%%%%%%%%%%%%%%%%%%%%%%%%%%

% %%%%%%%%%%%%%%%%%%%%%%%%%%%%%%%%%%%%%%%%%%%%%%%%%%%%%%%%%%%%%%%%%%%%
% %%%%%%%%%%%%%%%%%%%%%%%%%%%%%%%%%%%%%%%%%%%%%%%%%%%%%%%%%%%%%%%%%%%%

% %%%%%%%%%%%%%%%%%%%%%%%%%%%%%%%%%%%%%%%%%%%%%%%%%%%%%%%%%%%%%%%%%%%%
% %%%%%%%%%%%%%%%%%%%%%%%%%%%%%%%%%%%%%%%%%%

% %%%%%%%%%%%%%%%%%%%%%%%%%%%%%%%%%%%%%%%%%%
\begin{table*}%[t]

\begin{center}
\caption{Running time (sec.) comparison of different feature selection methods.}
\label{tab:running_time}
{\fontsize{10}{12}\selectfont\begin{tabular}{lccccc}
\toprule
                & ANOVA     & Feat. Imp.    & mRMR  & DFT (B=8)     & DFT (B=16) \\
\midrule
MultiFeat       & 0.011     & 363.39       & 15.19  & 2.74         & 5.78\\
Colon           & 0.003     & 58.99        & 23.15  & 1.55         & 3.19  \\ 
ARR             & 0.002     & 55.34        & 10.64  & 0.23         & 0.46  \\ 

\bottomrule
\end{tabular}}
\end{center}
\end{table*}
% %%%%%%%%%%%%%%%%%%%%%%%%%%%%%%%%%%%%%%%%%%

\subsubsection{DFT with feature pre-processing}

\begin{figure*}[tbp]
\centering
\begin{subfigure}{0.45\textwidth}
\centering
    \includegraphics[width=0.95\linewidth]{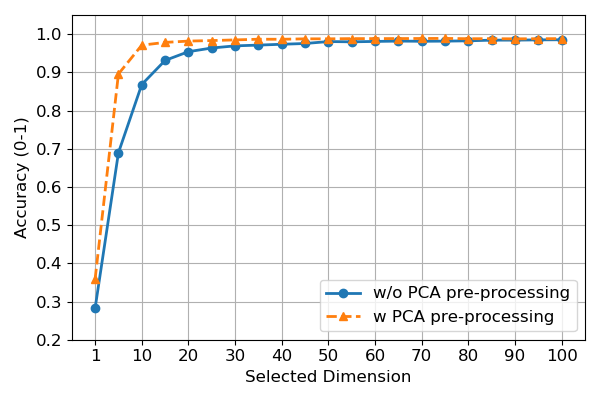}
    \caption{MNIST}
\end{subfigure}
\begin{subfigure}{0.45\textwidth}
\centering
    \includegraphics[width=0.95\linewidth]{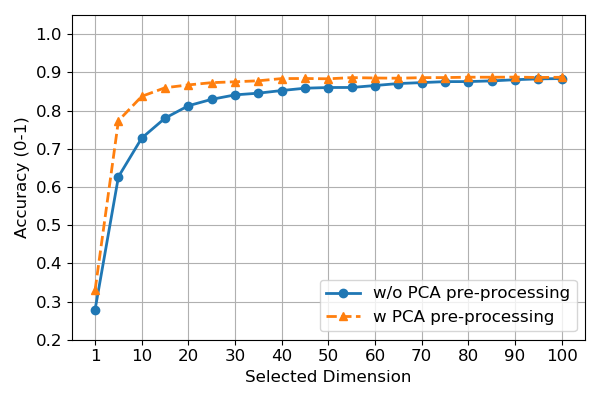}
    \caption{Fashion-MNIST}
\end{subfigure}
\vspace{-5pt}
\caption{Performance comparison of DFT feature selection with and without 
PCA feature pre-processing.}\label{fig:acc_comparison_PCA}
\end{figure*}

DFT assigns a score to each feature and selects a subset without any
pre-processing. Yet, there might be correlation between features so that
a redundant feature subset might be selected based on feature
ranking~\cite{chandrashekar2014survey}. Instead of adding redundancy
measure to the DFT loss, we study the effect of combining DFT with
feature pre-processing, such as PCA for feature decorrelation. We choose
clean MNIST and Fashion-MNIST datasets as examples and perform PCA on
the 400-D deep features without energy truncation. The DFT loss is
calculated for each of 400 PCA-decorrelated features.  After feature
selection, the XGBoost classifier is applied.
Fig.~\ref{fig:acc_comparison_PCA} compares the test accuracy under
different selected dimensions for each setting. We see that PCA pre-processing
improves the classification performance with the same selected dimension.

\begin{figure*}[htbp]
\centering
\begin{subfigure}{0.45\textwidth}
\centering
    \includegraphics[width=0.95\linewidth]{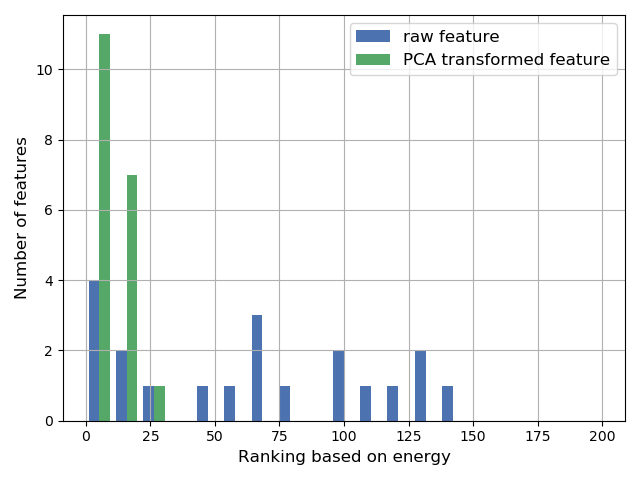}
    \caption{Select 20 features}
\end{subfigure}
\begin{subfigure}{0.45\textwidth}
\centering
    \includegraphics[width=0.95\linewidth]{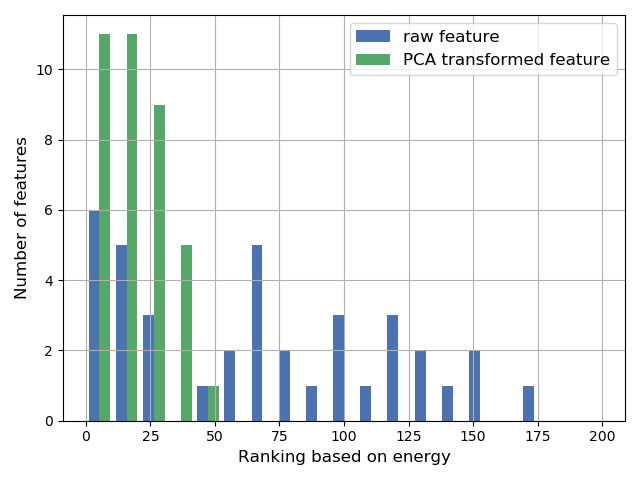}
    \caption{Select 40 features}
\end{subfigure}
\vspace{-5pt}
\caption{Histogram comparison of feature indices ranked by the energy
with 20 and 40 selected feature numbers before and after PCA
pre-processing. The smaller the ranking index in the x-axis, the higher
the feature energy.} \label{fig:PCA_selected_hist}
\end{figure*}

Furthermore, PCA pre-processing allows a smaller feature dimension
for the same performance. For example, the accuracy on Fashion-MNIST
saturates at around 15-D and 30-D with and without pre-processing,
respectively. This can be explained by the energy compaction capability
of PCA.  Fig.~\ref{fig:PCA_selected_hist} shows the histogram of energy
ranking of the feature subset selected by DFT with and without PCA
preprocessing.  The raw features are first sorted by decreasing energy
(variance) prior to feature selection. We see that the selected subset
tends to gather in the first 20 to 50 principal components with PCA
pre-preprocessing while the selected features are more widely
distributed without PCA.

% %%%%%%%%%%%%%%%%%%%%%%%%%%%%%%%%%%%%%%%%%%%%%%%%%%%%%%%%%%%%%%%%%%%

% %%%%%%%%%%%%%%%%%%%%%%%%%%%%%%%%%%%%%%%%%%%%%%%%%%%%%%%%%%%%%%%%%%%

% %%%%%%%%%%%%%%%%%%%%%%%%%%%%%%%%%%%%%%%%%%%%%%%%%%%%%%%%%%%%%%%%%%%
% %%%%%%%%%%%%%%%%%%%%%%%%%%%%%%%%%%%%%%%%%%%%%%%%%%%%%%%%%%%%%%%%%%%

\subsection{RFT for Regression Problems}

We convert the discrete class labels arising from the classification
problem to floating numbers so as to formulate a regression problem.  We
compare effectiveness of four feature selection methods: 1) variance
(Var.), 2) absolute correlation coefficient w.r.t the regression target
(Abs. Corr. Coeff.), 3) feature importance (Feat. Imp.) from a
pre-trained XGBoost regressor (of 50 trees), and 4) RFT. Again, we can
draw two conclusions. 

\subsubsection{RFT offers a more obvious elbow point}
\begin{figure*}[bp]
\centerline{\includegraphics[width=0.65\linewidth]{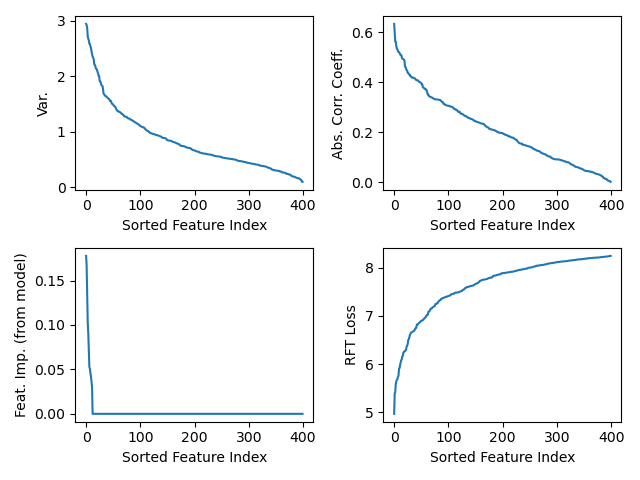}}
\vspace{-5pt}
\caption{Comparison of relevant feature selection capability among
four feature selection methods for the regression task on the 
Fashion-MNIST dataset.}
\label{fig:reg_fashion_4_featimp}
\end{figure*}

Fig.~\ref{fig:reg_fashion_4_featimp} compares the ranked scores for
different feature selection methods. The lower RFT loss, the higher
feature importance while the other three have a reversed relation.  RFT
has a more obvious elbow point than the knee points of Variance and
correlation-based methods.  The feature importance from the pre-trained
XGBoost regressor saturates very fast (up to 24-D) and the difference
among the remaining features is small. In contrast, RFT has a more
distinct and reasonable elbow point, ensuring the performance after
dimension reduction. A larger XGBoost model with more trees can help
increase the feature number of higher importance. Yet, it is not clear
what model size would be suitable for a particular regression problem. 
% %%%%%%%%%%%%%%%%%%%%%%%%%%%%%%%%%%%%%%%%%%%%%%%%%%%%%%%%%%%%%%%%%%%%
% %%%%%%%%%%%%%%%%%%%%%%%%%%%%%%%%%%%%%%%%%%%%%%%%%%%%%%%%%%%%%%%%%%%%

\subsubsection{Features selected by RFT achieves comparable and stable
performance} Table~\ref{tab:RFT_performance} summarizes the regression
MSE at two reduced dimensions selected by the RFT loss curves using
early and late elbow points. The proposed RFT can achieve comparable (or
even the best) performance among the four methods at the same selected
feature dimension regardless whether the input images are clean or
noisy. By employing only 25-37.5\% of the total feature dimensions, the
regression MSEs obtained by the late elbow point of RFT are 20-30\% and
5-10\% higher than those of the full feature set for MNIST and Fashion
MNIST, respectively. This demonstrates the effectiveness of the RFT
feature selection method. 

% %%%%%%%%%%%%%%%%%%%%%%%%%%%%%%%%%%%%%%%%%%%%%%%%%%%%%%%%%%%%%%%%%%%%
\begin{table*}[t]
    \small
	\centering
	\caption{Regression MSE comparison for MNIST (clean/noisy) images with features selected by four
		methods.}\label{tab:RFT_performance}
	{\fontsize{10}{12}\selectfont\begin{tabular}{@{}lccc@{}}
		\toprule
		\multirow{2}{*}{Method}         & \multicolumn{1}{l}{Early Elbow Point} & \multicolumn{1}{l}{Late Elbow Point} & Full Set                     \\
		& 30-D / 50-D                           & 100-D / 100-D                         & 400-D                                \\ 
		\midrule
		Var.                                                                                                        & 1.45 / \textbf{1.23}                  & \textbf{0.90} / \textbf{0.99}                           & \multirow{4}{*}{0.70 / 0.83}         \\
		Abs. Corr. Coeff.                                                                                                & \underline{1.43} / 1.37               & \textbf{0.90} / 1.06                           &                                      \\
		Feat. Imp.                                                                                                  & 1.55 / 1.47                           & 1.04 / 1.23                           &                                      \\
		RFT (Ours)                                                                                                  & \textbf{1.37} / \underline{1.36}      & \underline{0.91} / \underline{1.04}                           &                                      \\ 
		\bottomrule
	\end{tabular}}
	% \vspace{-5pt}
	% \end{adjustbox}
\end{table*}
% %%%%%%%%%%%%%%%%%%%%%%%%%%%%%%%%%%%%%%%%%%%%%%%%%%%%%%%%%%%%%%%%%%%%

% %%%%%%%%%%%%%%%%%%%%%%%%%%%%%%%%%%%%%%%%%%%%%%%%%%%%%%%%%%%%%%%%%%%%
\begin{table*}[t]
    \small
	\centering
	\caption{Regression MSE comparison for Fashion-MNIST (clean/noisy) images with features selected by four
		methods.}\label{tab:RFT_performance_fashion}
	{\fontsize{10}{12}\selectfont\begin{tabular}{@{}lccc@{}}
			\toprule
			\multirow{2}{*}{Method} & \multicolumn{1}{l}{Early Elbow Point} & \multicolumn{1}{l}{Late Elbow Point} & \multicolumn{1}{l}{Full Set} \\
			\multicolumn{1}{c}{}                                                                                       & 30-D / 50-D                           & 150-D / 150-D                         & 400-D                                \\ 
			
			\midrule
			Var.                                                                                                        & 2.08 / \underline{1.98}               & \textbf{1.46} / \textbf{1.73}                           & \multirow{4}{*}{1.35 / 1.62}         \\
			Abs. Corr. Coeff.                                                                                                & \textbf{1.95} / \textbf{1.96}         & 1.49 / 1.75                           &                                      \\
			Feat. Imp.                                                                                                  & 2.00 / 2.06                           & 1.62 / 1.86                           &                                      \\
			RFT (Ours)                                                                                                  & \underline{1.97} / \textbf{1.96}      & \underline{1.48} / \textbf{1.73}                           &                                      \\ 
			\bottomrule
	\end{tabular}}
	% \vspace{-5pt}
	% \end{adjustbox}
\end{table*}
% %%%%%%%%%%%%%%%%%%%%%%%%%%%%%%%%%%%%%%%%%%%%%%%%%%%%%%%%%%%%%%%%%%%%

\section{Conclusion and Future Work}\label{sec:conclusion}

Two feature selection methods, DFT and RFT, were proposed for general
classification and regression tasks in this work.  As compared with
other existing feature selection methods, DFT and RFT are effective in
finding distinct feature subspaces by offering obvious elbow regions in
DFT/RFT curves.  They provide feature subspaces of significantly lower
dimensions while maintaining near optimal classification/regression
performance. They are computationally efficient. They are
also robust to noisy input data.  

Recently, there is an emerging research direction that targets at
unsupervised representation learning \cite{chen2018saak,
chen2020pixelhop, chen2020pixelhop++, liu2021voxelhop,
manimaran2020visualization, zhang2020pointhop, zhang2020pointhop++}.
Through this process, it is easy to get high dimensional feature spaces
(say, 1000-D or higher). We plan to apply DFT/RFT to them and find
discriminant/relevant feature subspaces for specific tasks.

\section*{Acknowledgement}\label{sec:acknowledgement}

The authors acknowledge the Center for Advanced Research Computing
(CARC) at the University of Southern California for providing computing
resources that have contributed to the research results reported within
this publication. 

%\clearpage
\bibliographystyle{ieeetr}
\bibliography{ref}

\begin{thebibliography}{10}

\bibitem{hammer1962adaptive}
P.~Hammer, ``Adaptive control processes: a guided tour (r. bellman),'' 1962.

\bibitem{tang2014feature}
J.~Tang, S.~Alelyani, and H.~Liu, ``Feature selection for classification: A
  review,'' {\em Data classification: Algorithms and applications}, p.~37,
  2014.

\bibitem{miao2016survey}
J.~Miao and L.~Niu, ``A survey on feature selection,'' {\em Procedia Computer
  Science}, vol.~91, pp.~919--926, 2016.

\bibitem{venkatesh2019review}
B.~Venkatesh and J.~Anuradha, ``A review of feature selection and its
  methods,'' {\em Cybernetics and Information Technologies}, vol.~19, no.~1,
  pp.~3--26, 2019.

\bibitem{guyon2003introduction}
I.~Guyon and A.~Elisseeff, ``An introduction to variable and feature
  selection,'' {\em Journal of machine learning research}, vol.~3, no.~Mar,
  pp.~1157--1182, 2003.

\bibitem{chen2018saak}
Y.~Chen, Z.~Xu, S.~Cai, Y.~Lang, and C.-C.~J. Kuo, ``A saak transform approach
  to efficient, scalable and robust handwritten digits recognition,'' in {\em
  2018 Picture Coding Symposium (PCS)}, pp.~174--178, IEEE, 2018.

\bibitem{chen2020pixelhop}
Y.~Chen and C.-C.~J. Kuo, ``Pixelhop: A successive subspace learning (ssl)
  method for object recognition,'' {\em Journal of Visual Communication and
  Image Representation}, vol.~70, p.~102749, 2020.

\bibitem{chen2020pixelhop++}
Y.~Chen, M.~Rouhsedaghat, S.~You, R.~Rao, and C.-C.~J. Kuo, ``Pixelhop++: A
  small successive-subspace-learning-based (ssl-based) model for image
  classification,'' in {\em 2020 IEEE International Conference on Image
  Processing (ICIP)}, pp.~3294--3298, IEEE, 2020.

\bibitem{kuo2018data}
C.-C.~J. Kuo and Y.~Chen, ``On data-driven saak transform,'' {\em Journal of
  Visual Communication and Image Representation}, vol.~50, pp.~237--246, 2018.

\bibitem{kuo2019interpretable}
C.-C.~J. Kuo, M.~Zhang, S.~Li, J.~Duan, and Y.~Chen, ``Interpretable
  convolutional neural networks via feedforward design,'' {\em Journal of
  Visual Communication and Image Representation}, 2019.

\bibitem{liu2021voxelhop}
X.~Liu, F.~Xing, C.~Yang, C.-C.~J. Kuo, S.~Babu, G.~E. Fakhri, T.~Jenkins, and
  J.~Woo, ``Voxelhop: Successive subspace learning for als disease
  classification using structural mri,'' {\em arXiv preprint arXiv:2101.05131},
  2021.

\bibitem{manimaran2020visualization}
A.~Manimaran, T.~Ramanathan, S.~You, and C.-C.~J. Kuo, ``Visualization,
  discriminability and applications of interpretable saak features,'' {\em
  Journal of Visual Communication and Image Representation}, vol.~66,
  p.~102699, 2020.

\bibitem{rouhsedaghat2020facehop}
M.~Rouhsedaghat, Y.~Wang, X.~Ge, S.~Hu, S.~You, and C.-C.~J. Kuo, ``Facehop: A
  light-weight low-resolution face gender classification method,'' {\em arXiv
  preprint arXiv:2007.09510}, 2020.

\bibitem{zhang2020pointhop}
M.~Zhang, H.~You, P.~Kadam, S.~Liu, and C.-C.~J. Kuo, ``Pointhop: An
  explainable machine learning method for point cloud classification,'' {\em
  IEEE Transactions on Multimedia}, 2020.

\bibitem{zhang2020pointhop++}
M.~Zhang, Y.~Wang, P.~Kadam, S.~Liu, and C.-C.~J. Kuo, ``Pointhop++: A
  lightweight learning model on point sets for 3d classification,'' in {\em
  2020 IEEE International Conference on Image Processing (ICIP)},
  pp.~3319--3323, IEEE, 2020.

\bibitem{mitra2002unsupervised}
P.~Mitra, C.~Murthy, and S.~K. Pal, ``Unsupervised feature selection using
  feature similarity,'' {\em IEEE transactions on pattern analysis and machine
  intelligence}, vol.~24, no.~3, pp.~301--312, 2002.

\bibitem{cai2010unsupervised}
D.~Cai, C.~Zhang, and X.~He, ``Unsupervised feature selection for multi-cluster
  data,'' in {\em Proceedings of the 16th ACM SIGKDD international conference
  on Knowledge discovery and data mining}, pp.~333--342, 2010.

\bibitem{qian2013robust}
M.~Qian and C.~Zhai, ``Robust unsupervised feature selection,'' in {\em
  Twenty-third international joint conference on artificial intelligence},
  Citeseer, 2013.

\bibitem{solorio2020review}
S.~Solorio-Fern{\'a}ndez, J.~A. Carrasco-Ochoa, and J.~F.
  Mart{\'\i}nez-Trinidad, ``A review of unsupervised feature selection
  methods,'' {\em Artificial Intelligence Review}, vol.~53, no.~2,
  pp.~907--948, 2020.

\bibitem{sheikhpour2017survey}
R.~Sheikhpour, M.~A. Sarram, S.~Gharaghani, and M.~A.~Z. Chahooki, ``A survey
  on semi-supervised feature selection methods,'' {\em Pattern Recognition},
  vol.~64, pp.~141--158, 2017.

\bibitem{zhao2007semi}
Z.~Zhao and H.~Liu, ``Semi-supervised feature selection via spectral
  analysis,'' in {\em Proceedings of the 2007 SIAM international conference on
  data mining}, pp.~641--646, SIAM, 2007.

\bibitem{huang2015supervised}
S.~H. Huang, ``Supervised feature selection: A tutorial.,'' {\em Artif. Intell.
  Res.}, vol.~4, no.~2, pp.~22--37, 2015.

\bibitem{lee1999learning}
D.~D. Lee and H.~S. Seung, ``Learning the parts of objects by non-negative
  matrix factorization,'' {\em Nature}, vol.~401, no.~6755, pp.~788--791, 1999.

\bibitem{hartigan1972direct}
J.~A. Hartigan, ``Direct clustering of a data matrix,'' {\em Journal of the
  american statistical association}, vol.~67, no.~337, pp.~123--129, 1972.

\bibitem{aggarwal1999fast}
C.~C. Aggarwal, J.~L. Wolf, P.~S. Yu, C.~Procopiuc, and J.~S. Park, ``Fast
  algorithms for projected clustering,'' {\em ACM SIGMoD Record}, vol.~28,
  no.~2, pp.~61--72, 1999.

\bibitem{kohavi1997wrappers}
R.~Kohavi and G.~H. John, ``Wrappers for feature subset selection,'' {\em
  Artificial intelligence}, vol.~97, no.~1-2, pp.~273--324, 1997.

\bibitem{rfe}
I.~Guyon, J.~Weston, S.~Barnhill, and V.~Vapnik, ``Gene selection for cancer
  classification using support vector machines,'' {\em Machine learning},
  vol.~46, no.~1, pp.~389--422, 2002.

\bibitem{anova}
H.~Scheffe, {\em The analysis of variance}, vol.~72.
\newblock John Wiley \& Sons, 1999.

\bibitem{xgb}
T.~Chen, T.~He, M.~Benesty, V.~Khotilovich, Y.~Tang, H.~Cho, K.~Chen, {\em
  et~al.}, ``Xgboost: extreme gradient boosting,'' {\em R package version
  0.4-2}, vol.~1, no.~4, pp.~1--4, 2015.

\bibitem{peng2005feature}
H.~Peng, F.~Long, and C.~Ding, ``Feature selection based on mutual information
  criteria of max-dependency, max-relevance, and min-redundancy,'' {\em IEEE
  Transactions on pattern analysis and machine intelligence}, vol.~27, no.~8,
  pp.~1226--1238, 2005.

\bibitem{ding2005minimum}
C.~Ding and H.~Peng, ``Minimum redundancy feature selection from microarray
  gene expression data,'' {\em Journal of bioinformatics and computational
  biology}, vol.~3, no.~02, pp.~185--205, 2005.

\bibitem{lecun1998gradient}
Y.~LeCun, L.~Bottou, Y.~Bengio, and P.~Haffner, ``Gradient-based learning
  applied to document recognition,'' {\em Proceedings of the IEEE}, vol.~86,
  no.~11, pp.~2278--2324, 1998.

\bibitem{xiao2017fashion}
H.~Xiao, K.~Rasul, and R.~Vollgraf, ``Fashion-mnist: a novel image dataset for
  benchmarking machine learning algorithms,'' {\em arXiv preprint
  arXiv:1708.07747}, 2017.

\bibitem{guvenir1997supervised}
H.~A. Guvenir, B.~Acar, G.~Demiroz, and A.~Cekin, ``A supervised machine
  learning algorithm for arrhythmia analysis,'' in {\em Computers in Cardiology
  1997}, pp.~433--436, IEEE, 1997.

\bibitem{Dua:2019}
D.~Dua and C.~Graff, ``{UCI} machine learning repository,'' 2017.

\bibitem{alon1999broad}
U.~Alon, N.~Barkai, D.~A. Notterman, K.~Gish, S.~Ybarra, D.~Mack, and A.~J.
  Levine, ``Broad patterns of gene expression revealed by clustering analysis
  of tumor and normal colon tissues probed by oligonucleotide arrays,'' {\em
  Proceedings of the National Academy of Sciences}, vol.~96, no.~12,
  pp.~6745--6750, 1999.

\bibitem{van1998handwritten}
M.~van Breukelen, R.~P. Duin, D.~M. Tax, and J.~Den~Hartog, ``Handwritten digit
  recognition by combined classifiers,'' {\em Kybernetika}, vol.~34, no.~4,
  pp.~381--386, 1998.

\bibitem{van1998neural}
M.~Van~Breukelen and R.~P. Duin, ``Neural network initialization by combined
  classifiers,'' in {\em Proceedings. Fourteenth International Conference on
  Pattern Recognition (Cat. No. 98EX170)}, vol.~1, pp.~215--218, IEEE, 1998.

\bibitem{jain1997feature}
A.~Jain and D.~Zongker, ``Feature selection: Evaluation, application, and small
  sample performance,'' {\em IEEE transactions on pattern analysis and machine
  intelligence}, vol.~19, no.~2, pp.~153--158, 1997.

\bibitem{logisticregression}
D.~R. Cox, ``The regression analysis of binary sequences,'' {\em Journal of the
  Royal Statistical Society: Series B (Methodological)}, vol.~20, no.~2,
  pp.~215--232, 1958.

\bibitem{svm}
C.~Cortes and V.~Vapnik, ``Support-vector networks,'' {\em Machine learning},
  vol.~20, no.~3, pp.~273--297, 1995.

\bibitem{RF}
L.~Breiman, ``Random forests,'' {\em Machine learning}, vol.~45, no.~1,
  pp.~5--32, 2001.

\bibitem{chandrashekar2014survey}
G.~Chandrashekar and F.~Sahin, ``A survey on feature selection methods,'' {\em
  Computers \& Electrical Engineering}, vol.~40, no.~1, pp.~16--28, 2014.

\end{thebibliography}

\end{document}